\documentclass[runningheads]{llncs}
\usepackage{graphicx}
\usepackage{comment}
\usepackage{amsmath,amssymb} 
\usepackage{color}
\usepackage{multirow}
\usepackage{bm}

\makeatletter
\newcommand{\printfnsymbol}[1]{%
  \textsuperscript{\@fnsymbol{#1}}%
}
\makeatletter

\newcommand{\Precision}{\mathrm{Precision}}
\newcommand{\Recall}{\mathrm{Recall}}
\newcommand{\IoU}{\mathrm{IoU}}
\newcommand{\Score}{\mathrm{Score}}

\newcommand{\tabincell}[2]{\begin{tabular}{@{}#1@{}}#2\end{tabular}}

\begin{document}
\pagestyle{headings}
\mainmatter
\def\ECCVSubNumber{2709}

\newcommand{\hy}[1]{\textcolor{red}{#1}}
\newcommand{\junz}[1]{\textcolor{blue}{[zj: #1]}}

\title{Design and Interpretation of Universal Adversarial Patches in Face Detection} 

\titlerunning{Adversarial Patches in Face Detection}
%
\author{Xiao Yang$^{1}$\thanks{Equal contribution. $^\ddag$ corresponding author.} ~~ Fangyun Wei $^{2}$\printfnsymbol{1} ~~  Hongyang Zhang$^{3}$\printfnsymbol{1} ~~ Jun Zhu$^{1}$$^\ddag$}
\authorrunning{X.Yang et al.}
%
\institute{Dept. of Comp. Sci. \& Tech.,  BNRist Center, Institute for AI, Tsinghua University$^{1}$ \\
	Microsoft Research Asia$^{2}$ \qquad TTIC$^{3}$\\
\email{ \{yangxiao19@mails, dcszj@\}.tsinghua.edu.cn$^{1}$  \\
	\tt\small fawe@microsoft.com$^{2}$ \qquad hongyanz@ttic.edu$^{3}$ }}
\maketitle

\begin{abstract}
We consider universal adversarial patches for faces --- small visual elements whose addition to a face image reliably destroys the performance of face detectors. Unlike previous work that mostly focused on the algorithmic design of adversarial examples in terms of improving the success rate as an attacker, in this work we show an \emph{interpretation} of such patches that can prevent the state-of-the-art face detectors from detecting the real faces.
We investigate a phenomenon: patches designed to suppress real face detection appear face-like. This phenomenon holds generally across different initialization, locations, scales of patches, backbones, and state-of-the-art face detection frameworks. We propose new optimization-based approaches to automatic design of universal adversarial patches for varying goals of the attack, including scenarios in which true positives are suppressed without introducing false positives. Our proposed algorithms perform well on real-world datasets, deceiving state-of-the-art face detectors in terms of multiple precision/recall metrics and transferability.

\vspace{-0.3cm}
\end{abstract}

\vspace{-0.7cm}
\section{Introduction}
\vspace{-0.1cm}
Adversarial examples still remain knotty in computer vision~\cite{brendel2020adversarial,xie2017adversarial,dong2020benchmarking}, machine learning~\cite{zhang2019theoretically,madry2017towards}, security~\cite{PapernotDistillation2016}, and other domains~\cite{jia2017adversarial} despite the huge success of deep neural networks~\cite{you2020greedynas,you2017learning,wei2020point}. In computer vision and machine learning, study of adversarial examples serves as evidences of substantial discrepancy between the human vision system and machine perception mechanism~\cite{szegedy2013intriguing,Nguyen2015,biggio2013evasion,goodfellow2014explaining}. In security, adversarial examples have raised major concerns on the vulnerability of machine learning systems to malicious attacks. The problem can be stated as modifying an image, subject to some constraints, so that learning system's response is drastically altered, e.g., changing the classifier or detector output from correct to incorrect. The constraints either come in the \emph{human-imperceptible} form such as bounded $\ell_p$ perturbations~\cite{zhang2019theoretically,yang2020adversarial,blum2020random}, or in the \emph{human-perceptible} form such as small patches~\cite{thys2019fooling,eykholt2018robust}. The focus of this work is the latter setting.

While image classification has been repeatedly shown to be broadly vulnerable to adversarial attacks~\cite{szegedy2013intriguing}, it is less clear whether object detection is similarly vulnerable~\cite{ren2015faster,liu2016ssd,lin2017feature,lin2017focal,he2017mask}. State-of-the-art detectors propose thousands of candidate bounding boxes and the adversarial examples are required to fool all of them simultaneously. Nonetheless, for selected object categories the attacks and defenses have been studied extensively. They include objects like stop signs or pedestrians~\cite{liu2018dpatch,thys2019fooling,eykholt2018robust}, but few attempts have been made on generating adversarial examples for faces.
This is in spite of face detection as a task enjoying significant attention in recent years, due to its practical significance on its own and as a building block for applications such as face alignment, recognition, attribute analysis, and tracking. Publicly available face detectors~\cite{zhu2018seeing,ming2019group,li2019dsfd,zhang2019single} can achieve performance on par with humans, e.g., on FDDB~\cite{jain2010fddb} and WIDER FACE dataset~\cite{yang2016wider}, and are insensitive to the variability in occlusions, scales, poses and lighting. However, much remains unknown concerning the behaviors of face detectors on adversarial patches. Our work sheds new light on this question and shows that a simple approach of pasting a single universal patch onto a face image can dramatically harm the accuracy of state-of-the-art face detectors. We propose multiple approaches for building adversarial patches, that address different desired precision/recall characteristics of the resulting performance. In addition to empirical performance, we are interested in understanding the nature of adversarial patch on face detection.

\begin{figure*}[t]
\centering
\includegraphics[width=0.9\linewidth]{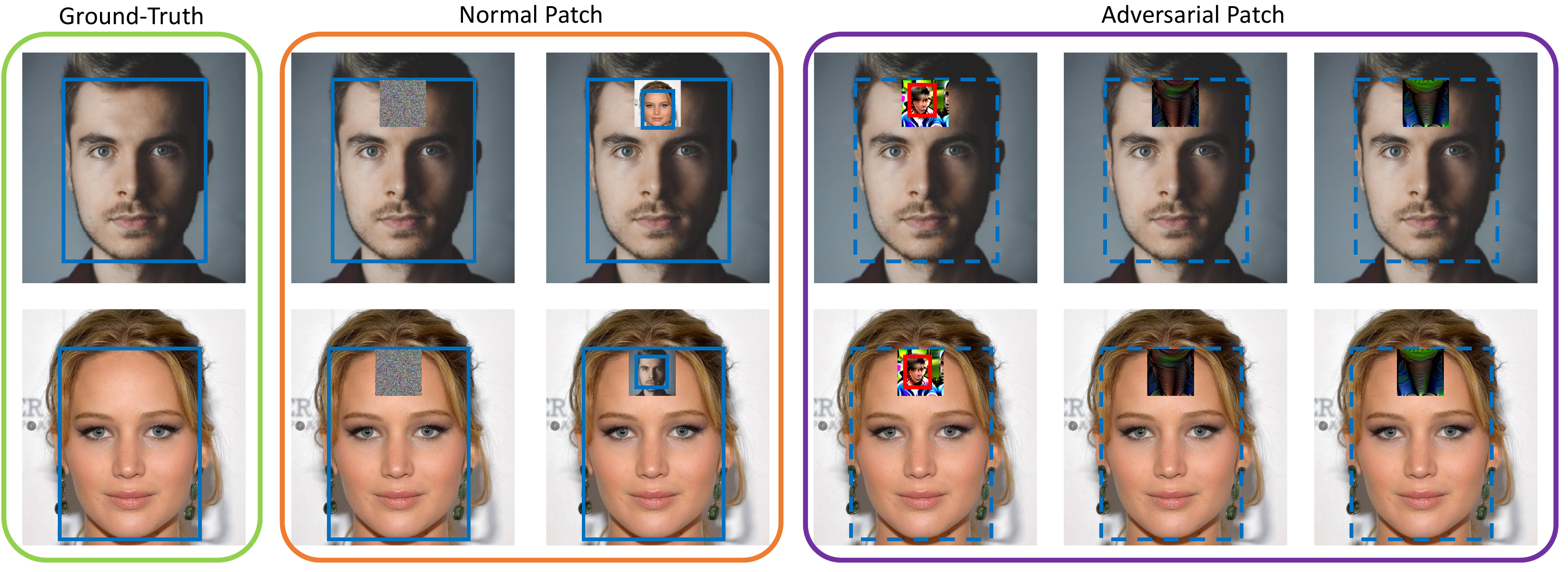}
\vspace{-2ex}
\caption{Properties and effect of different patches. In each image we show true positive (solid blue lines), false positive (red) and missed detection (dashed blue lines). \textbf{Left (green) box:} the clean input images. \textbf{Middle (orange) box:} pasting an un-optimized noise patch or a downsized face patch on the image does not affect the detectors. \textbf{Right (purple) box:} universal adversarial patches produced by different methods successfully suppress true positives, From the fourth to sixth column: The Baseline method \emph{Patch-IoU} appears person-like and induces false positives; our \textit{Patch}-\textit{Score}-\textit{Focal} and \textit{Patch}-\textit{Combination} avoid the false positives. The patches are not necessarily pasted at forehead as demonstrated in Section~\ref{sec:interpretation}.}
\label{fig:adv-figure}  
\vspace{-0.7cm}
\end{figure*}

\noindent{\textbf{Significance.}}
The study of adversarial patch in face detection is important in multiple aspects: a) In security, adversarial patch serves as one of the most common forms of physical attacks in the detection problems, among which face detection has received significant attention in recent years. b) The study of adversarial patch may help understand the discrepancy between state-of-the-art face detectors and human visual system, towards algorithmic designs of detection mechanism as robust as humans. c) Adversarial patch to face detectors is human-perceptible and demonstrates significant interpretation as shown in this paper.

\noindent{\textbf{Challenges.}}
In commonly studied classification problems, adversarial perturbations are inscrutable and appear to be unstructured, random noise-like. Even when structure is perceptible, it tends to bear no resemblance to the categories involved.
Many observations and techniques for classification break down when we consider more sophisticated face detection tasks.
Compared with other detection tasks, generating adversarial examples for face detection is more challenging, because the state-of-the-art face detectors are able to detect very small faces (e.g., $6\times 6$ pixels \cite{ming2019group}) by applying the multi-scale training and testing data augmentation. While there is a large literature on the algorithmic designs of adversarial examples in terms of improving the success rate as an attacker, in this work we focus on the \emph{interpretation} of learning a small, universal adversarial patch which, once being attached to human faces, can prevent the state-of-the-art face detectors from detecting the real faces.

\begin{figure*}[t]
\begin{center}
\vspace{-2ex}
\includegraphics[width=0.9\linewidth]{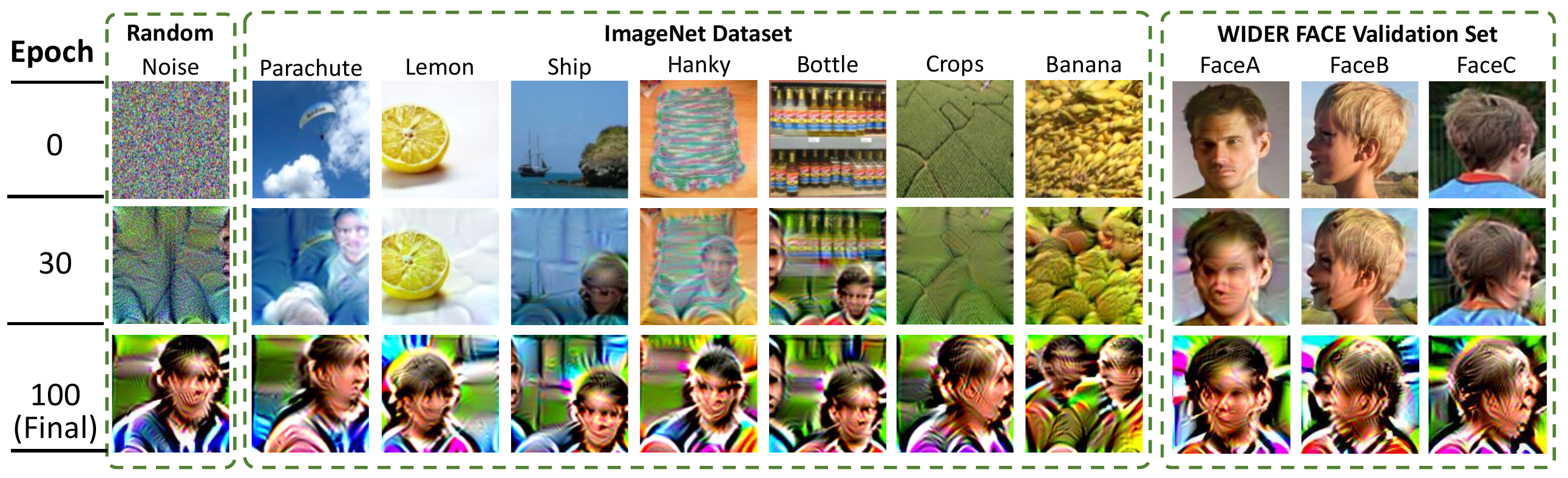}
\end{center}
\vspace{-5ex}
\caption{Adversarial patches from different \textit{initialization} by \textit{Patch-IoU}. The first row is the initial patches, and the second row and the third row represent the intermediate and final patches, respectively. All of the final patches here are detected as faces by face detectors.}
\label{fig:noise-init}
\vspace{-4ex}
\end{figure*}

\noindent{\textbf{Our results.}}
The gist of our findings is summarized in Figure~\ref{fig:adv-figure}. We consider state-of-the-art face detectors, that perform very accurately on natural face images. We optimize a universal adversarial patch, to be pasted on input face images, with the objective of suppressing scores of true positive detection on training data. This is in sharp contrast to most of existing works on adversarial examples for faces in the form of sample-specific, imperceptible perturbations, but a \emph{universal} (independent of the input image) and \emph{interpretable} (semantically meaningful) patch that reliably destroys the performance of face detectors is rarely studied in the literature. Our resulting patch yields the following observations.
\vspace{-0.2cm}
\begin{itemize}
\setlength{\itemsep}{2pt}
\setlength{\parsep}{0pt}
\setlength{\parskip}{0pt}
\item 
It succeeds in drastically suppressing true positives in test data. The attack also transfers between different face detection frameworks, that is, a patch which is trained on one detection framework deceives another detection framework with a high success rate.
\item
It looks face-like to humans, as well as to the detectors. Thus, in addition to reducing recall, it reduces precision by inducing false positives.
\item
Despite superficial face-likeness of the learned adversarial patch, it cannot be simply replaced by a real face patch, nor by a random noise pattern; affixing these to real faces does not fool the detectors.
\item
Surprisingly, these observations hold generally across different detection frameworks, patch initialization, locations and scales of pasted patch, etc. For example, even while initializing the patch with an image of a non-face object or a complex scene, after 100 epochs  the resulting adversarial patch comes to resemble a face (see Figure \ref{fig:noise-init}).
\end{itemize}
In some scenarios the attacker may want to suppress correct detection without creating false positives (e.g., to hide the presence of any face). We propose modified approaches that produce patches with this property. Intuitively, the approaches minimize the confidence scores of bounding boxes as long as they are larger than a threshold. Experiments verify the effectiveness of the proposed approaches (see the last two columns in Figure \ref{fig:adv-figure}).

\noindent{\textbf{Summary of contributions.}} Our work explores the choices in design of universal adversarial patches for face detection.
\begin{itemize}
\item
We show how such patches can be optimized to harm performance of existing face detectors. We also show that when the objective is purely to suppress true detection, the resulting patches are interpretable as face-like and can be detected by baseline detectors, with this property holding true across different experimental settings.
\item
In response to some security-focused scenarios where the adversary may want to suppress correct detection without creating false positives, we describe methods to produce equally successful universal adversarial patches that do not look like faces to either humans nor the face detectors, thus reducing detection rate without increasing false positives. Our proposed algorithms deceive the state-of-the-art face detectors~\cite{ming2019group} on real-world datasets in terms of multiple precision/recall metrics and transferability.
\end{itemize}

\vspace{-0.5cm}
\section{Related Work}
\vspace{-0.2cm}
\noindent{\textbf{Adversarial examples on object detection.}} Adversarial examples on general object detection have been extensively studied in the recent years~\cite{zhang2019towards,lee2019physical}. A commonly explored domain for adversarial examples in detection is stop sign detection~\cite{eykholt2018physical,eykholt2018robust,eykholt2017note,chen2018shapeshifter}. 
Inspired by an observation that both segmentation and detection are based on classifying multiple targets on an image, \cite{xie2017adversarial} extended the methodology of generating adversarial examples to the general object detection tasks. Recently, \cite{thys2019fooling} proposed a method to generate a universal adversarial patch to fool YOLO detectors on pedestrian data set. Another line of research related to our work is the perturbation-based adversarial examples for face detectors~\cite{li2019hiding}. This line of works adds sample-specific, human-imperceptible perturbations to the images globally. In contrast, our adversarial patches are universal to all samples, and our patches are visible to humans and show strong interpretation. While optimizing a patch to fool detectors has previously been used as a simulation of physical-world attacks, to our knowledge no properties of such patches to human visual system have been shown.

\noindent{\textbf{Adversarial examples in face recognition.}} To fool a face recognition system in the physical world, prior work has relied on active explorations via various forms of physical attacks~\cite{Kurakin2016}. For example, \cite{sharif2019general,sharif2016accessorize,yamada2013privacy} designed a pair of eyeglass frames which allows a face to evade being recognized or to impersonate another individual. Other attacks in the physical world that may fool (face) classifiers include adversarial patches~\cite{brown2017adversarial}, hats~\cite{komkov2019advhat}, and 3D-printing toys~\cite{athalye2017synthesizing}. However, these adversarial examples did not afford any semantic interpretation. Though scaled adversarial perturbations of robustly trained classifiers might have semantic meaning to humans~\cite{zhang2019theoretically}, those adversarially trained classifiers are not widely used due to an intrinsic trade-off between robustness and accuracy~\cite{tsipras2018robustness,zhang2019theoretically}. 

\noindent{\textbf{Face detection.}}
Face detection is typically less sensitive to the variation of face scales, angles, and other external factors such as occlusions and image qualities. Modern face detection algorithms~\cite{zhu2018seeing,ming2019group,li2019dsfd,zhang2019single} take advantage of anchor\footnote{Anchors are a set of predefined and well-designed initial rectangles with different scales and ratios. They are densely tiled on feature maps for object classification and bounding box regression.} based object detection methods, such as SSD~\cite{liu2016ssd} and RetinaNet~\cite{lin2017focal}, Faster R-CNN~\cite{ren2015faster} and Mask R-CNN~\cite{he2017mask}, and can achieve performance on par with humans on many public face detection benchmarks, such as FDDB and WIDER FACE dataset. They can detect faces as small as 6 pixels by applying multi-scale training and testing data augmentation, which serves as one of the primary differences with general object detection. Anchor based face detectors use IoU between anchors (or proposals) and ground truth to distinguish positive samples from negative ones in the sampling step during training. In the inference stage, average precision (\textit{AP}) is a commonly used metric to evaluate the detection performance. However, as we illustrate in the following sections, this criterion is not a fair metric to evaluate the impact of adversarial patches on a face detector.

\vspace{-0.3cm}
\section{Interpretation of Adversarial Patch as Face}
\vspace{-0.1cm}
\label{sec:interpretation}
In this section, we present our main experimental results on the interpretation of adversarial patch. We show that, on one hand, the adversarial patch optimized by the proposed \emph{Patch-IoU} method looks like a face. The patch can be detected by the baseline face detection model, even in the absence of extra constraints to encourage the patch to be face-like. On the other hand, attaching a face picture to a real face does not fool the detector (see Figure \ref{fig:adv-figure}). The phenomenon holds generally across different setups.

\vspace{-0.1cm}
\subsection{Preliminaries on face detection}
\vspace{-0.1cm}

\noindent{\textbf{Dataset.}}
We use WIDER FACE~\cite{yang2016wider} training dataset to learn both the face detector and the adversarial patch. The WIDER FACE dataset contains 32,203 images and 393,703 annotated face bounding boxes with high degree of variability in scales, poses, occlusions, expression, makeup, and illumination. According to the detection rate of EdgeBox~\cite{zitnick2014edge}, WIDER FACE dataset is split into 3 subsets: Easy, Medium and Hard. The face detector and adversarial patch are evaluated on the validation set. The set of ground-truth bounding boxes for an image is defined as $\{B_{i}\}$, where $B_{i} = (x^{i}, y^{i}, w^{i}, h^{i})$, $(x^{i}, y^{i})$ is the center of the box, and $w^{i}$ and $h^{i}$ are the width and height of the bounding box, respectively.
\begin{figure}[t]
\begin{center}
\includegraphics[width=0.68\linewidth]{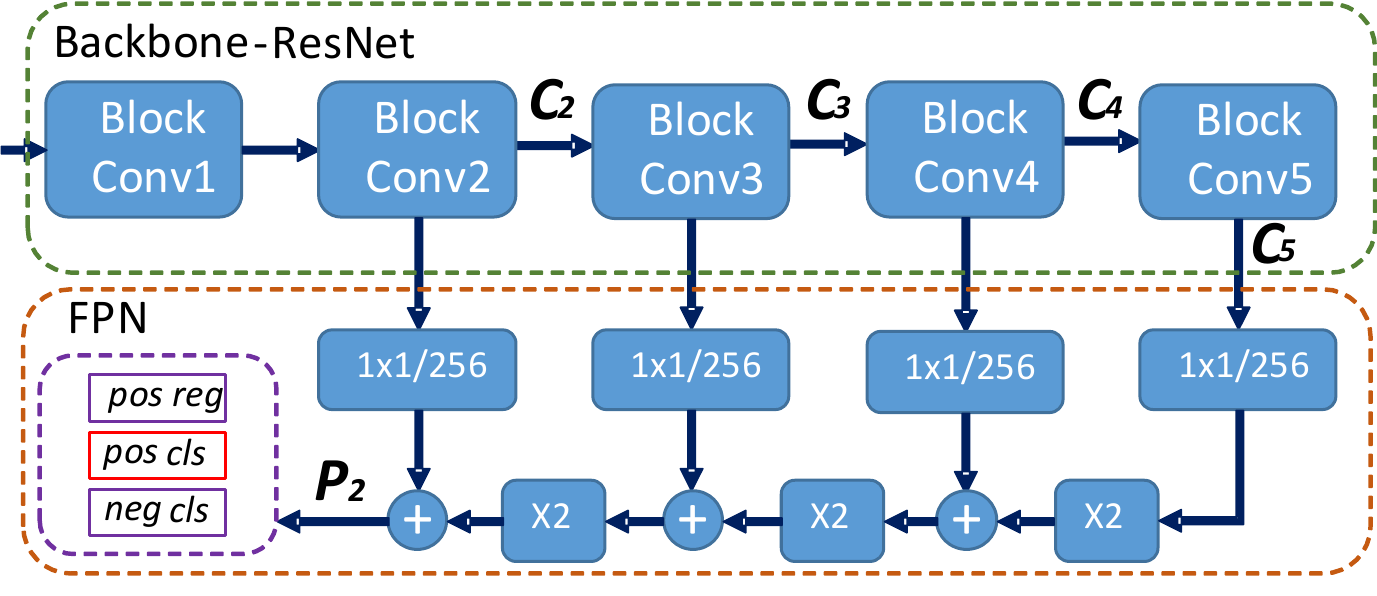}
\end{center}
\vspace{-6ex}
\caption{SLN framework in our face detection baseline model, where $\times2$ represents the bilinear upsampling, $+$ represents the element-wise summation, and $1\times1/256$ represents the $1\times1$ convolution with 256 output channels. The feature map $P_2$ is used as the only detection layer where all anchors are tiled with stride 4 pixels. The positive classification loss is our main attacking target.} 
\label{fig:resnet}
\vspace{-0.7cm}
\end{figure}

\noindent{\textbf{Face detection framework.}}
We use the state-of-the-art face detection framework~\cite{ming2019group} as the baseline model, and we name it as Single Level Network (SLN). Figure~\ref{fig:resnet} illustrates the network structure. We use ResNet~\cite{he2016deep} as backbone, with a bottom-up feature refusion procedure in Feature Pyramid Network (FPN)~\cite{lin2017feature}. We obtain a high-resolution and informative feature map $P_2$ (stride equals to 4 pixels). Anchors with scales $\{16, 32, 64, 128\}$ and aspect ratio 1 are tiled on $P_2$ uniformly. We denote by $\{A_i\}$ the set of all anchors. We apply IoU regression loss, anchor matching criterion, and group sampling strategy in \cite{ming2019group} to train our baseline model. Formally, let $\IoU(A_i, B_j)$ denote the IoU between the $i$-th anchor and the $j$-th ground-truth bounding box. Anchors with $\IoU(A_i, B_j)>0.6$ and $\IoU(A_i, B_j)<0.4$ will be set as positive and negative samples. Finally, we define a multi-task loss $\mathcal{L} = \mathcal{L}_{cls}^p + \mathcal{L}_{cls}^n + \mathcal{L}_{reg}$, where $\mathcal{L}_{cls}^p$ and $\mathcal{L}_{cls}^n$ denote the standard cross entropy loss for positive and negative samples, respectively, and $\mathcal{L}_{reg} = \frac{1}{N_{reg}} \sum_{(A_i, B_j)}\left\|1 - \IoU(A_i, B_j)\right\|_2^2$ represents the IoU least square regression loss. If not specified, we use ResNet-18 as our defaulted backbone.

\noindent{\textbf{Training details of face detector.}}
We use random horizontal flip and scale jittering as data augmentation during training. For scale jittering, each image is resized by a factor of $0.25 \times n$, where $n$ is randomly chosen from $[1, 8]$. Then a random cropping procedure is used to crop a patch from the resized image to ensure that the longer side of the image does not exceed 1,200 pixels. We set the initial learning rate as 0.01 and decay the learning rate by a factor of 0.1 on the 60-th and the 80-th epochs. The model is trained for 100 epochs with synchronized stochastic gradient descent over 8 NVIDIA Tesla P100 GPUs and 8 images per mini-batch (1 image per GPU). The momentum is $0.9$ and weight decay is set to be $5 \times 10^{-5}$. Backbone is initialized with ImageNet pre-trained weights. We fine-tune the model on the WIDER FACE training set and test on validation set with same image pyramid strategy as in the training procedure. We use Non-Maximum Suppression (NMS) as post-processing. The first line in Table~\ref{tab:base-table} shows the precision and recall of the baseline model. \textit{Easy}, \textit{Medium}, \textit{Hard}, and \textit{All}\footnote{Official WIDER FACE testing script (\url{http://shuoyang1213.me/WIDERFACE/}) only gives results of \textit{Easy}, \textit{Medium} and \textit{Hard} subsets. We reimplement the test script to support testing on the whole validation set.} represent the results from easy subset, medium subset, hard subset and the whole validation set, respectively.

\begin{table}[t]
\scriptsize
\setlength{\tabcolsep}{2mm}
  \caption{Precision and recall of SLN baseline model and pasting various patches with (without) \emph{Patch-IoU} algorithm on WIDER FACE validation set under $\delta = 0.99$ (see Figure \ref{fig:noise-init} for visualized results).}
  \label{tab:base-table}
  \centering
  \vspace{-2ex}
  \begin{tabular}{l|c|c|c|c}
    \hline
   Precision/ Recall & \textit{Easy} & \textit{Medium} & \textit{Hard} & \textit{All} \\
    \hline\hline
    Baseline-SLN & 99.0/ 73.4 & 99.4/ 62.4 & 99.4/ 27.9 & 99.4/ 22.5 \\
    \hline
  Noise w/o \textit{Patch}-\textit{IoU} & 99.1/ 54.9 & 99.4/ 41.5 & 99.4/ 17.6 & 99.4/ 14.2 \\
    Parachute w/o \textit{Patch}-\textit{IoU} & 99.1/ 51.8 & 99.3/ 37.2 & 99.3/ 15.8 & 99.3/ 12.8 \\
    Lemon w/o \textit{Patch}-\textit{IoU} & 98.9/ 53.4 & 99.2/ 39.4 & 99.2/ 16.7 & 99.2/ 13.4 \\
    Bottle w/o \textit{Patch}-\textit{IoU} & 99.1/ 53.5 & 99.4/ 41.1 & 99.4/ 17.3 & 99.4/ 13.9 \\
    Banana w/o \textit{Patch}-\textit{IoU} & 99.1/ 55.2 & 99.4/ 41.4 & 99.4/ 17.5 & 99.4/ 14.1 \\
    FaceA w/o \textit{Patch}-\textit{IoU} & 51.8/ 30.2 & 61.4/ 24.2 & 61.8/ 10.3 & 61.8/ 8.3 \\
    FaceB w/o \textit{Patch}-\textit{IoU} & 77.8/ 39.5 & 83.5/ 30.1 & 83.6/ 12.9 & 83.6/ 10.4 \\
    FaceC w/o \textit{Patch}-\textit{IoU} & 98.4/ 38.3 & 98.9/ 29.8 & 98.9/ 12.7 & 98.9/ 10.2 \\
    \hline
    Noise w/ \textit{Patch}-\textit{IoU} & 2.7/2.7 & 6.5/3.7 & 7.3/1.8 & 7.3/1.4 \\
    Parachute w/ \textit{Patch}-\textit{IoU} & 2.1/0.5 & 4.8/ 0.7 & 5.9/ 0.4 & 5.9/ 0.3 \\
    Lemon w/ \textit{Patch}-\textit{IoU} & 0.2/0.1 & 0.9/0.3 & 1.0/0.2 & 1.0/0.1 \\
    Bottle w/ \textit{Patch}-\textit{IoU} & 1.1/1.1 & 2.2/1.2 & 2.5/ 0.6 & 2.6/0.5 \\
    Banana w/ \textit{Patch}-\textit{IoU} & 10.2/5.0 & 19.0/5.6 & 20.3/2.5 & 20.3/2.0 \\
    FaceA w/ \textit{Patch}-\textit{IoU} & 0.0/0.0 & 0.0/0.0 & 0.0/0.0 & 0.0/0.0 \\
    FaceB w/ \textit{Patch}-\textit{IoU} & 0.1/0.0 & 2.3/0.2 & 2.6/0.1 & 2.6/0.1 \\
    FaceC w/ \textit{Patch}-\textit{IoU} & 0.1/0.1 & 0.2/0.1 & 0.3/0.0 & 0.3/0.0 \\
    \hline
  \end{tabular}
  \vspace{-5ex}
\end{table}

\vspace{-0.1cm}
\subsection{Design of adversarial patch}
\vspace{-0.1cm}

\label{basic_attack}
\noindent{\textbf{Details of adversarial patch.}}
In our work we need craft a universal adversarial patch that can be pasted at input faces, with the objective of fooling face detectors. For detection, as opposed to balanced classification problems, there are two types of errors: the false-positive error and the false-negative error.
In response to an intrinsic trade-off between precision and recall in the face detection tasks, existing works set a score threshold $\delta$ to keep high precision: output proposals with confidence scores higher than $\delta$ are treated as faces. The goal of adversary is to decrease the confidence scores to be lower than $\delta$ by pasting a carefully-calculated, universal patch to human faces. We show that the adversarial patch can make the real faces invisible to various detectors.
Formally, we define the patch $P$ as a rectangle which is denoted by $(x^{P}, y^{P}, w^{P}, h^{P} )$, where $(x^{P}, y^{P})$ is the center of the patch relative to the ground-truth bounding box $B_i$, and $w^{P}$ and $h^{P}$ represent its width and height, respectively. In our experiments, we set both $w^{P}$ and $h^{P}$ as 128 since the largest anchor size is 128 in the SLN face detection framework. For each of the ground-truth bounding box $B_i = (x^i, y^i, w^i, h^i)$ in the given training image, the patch $P$ is resized to $\alpha\sqrt{w^ih^i}$ ($0<\alpha<1$) and then placed on $B_i$ with its center position $(x^{P}, y^{P})$. We randomly initialize the patch, and set $\alpha = 0.5$ and $(x^{P}, y^{P}) = (w^i/2, \alpha\sqrt{w^ih^i}/2)$, unless otherwise specified. All of the training settings, including the training dataset and the hyper-parameter tuning, are the same as the SLN (or other face detection framework) baseline model.
\begin{figure}[t]
\begin{center}
\includegraphics[width=0.9\linewidth]{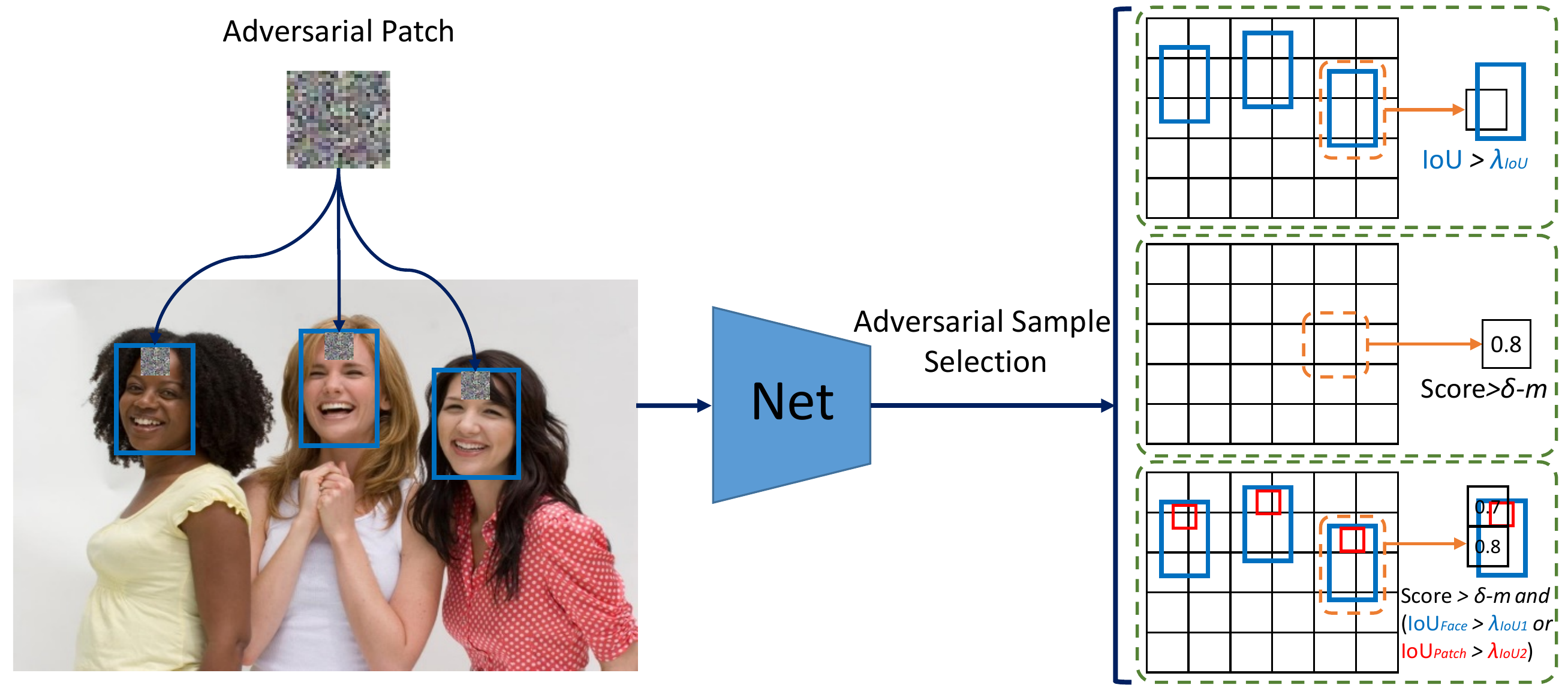}
\end{center}
\vspace{-5ex}
\caption{Different optimization methods for generating adversarial patches. The core difference is to involve different samples in the optimization of adversarial patch. From top to bottom: \textit{Patch}-\textit{IoU}, \textit{Patch}-\textit{Score}-\textit{Focal} and \textit{Patch}-\textit{Combination}, respectively.}
\label{fig:algo}
\vspace{-4ex}
\end{figure}

\noindent{\textbf{Optimization.}}
Previous adversarial attacks on object detectors~\cite{liu2018dpatch} have shown some progress through inverse optimization for the loss function of the detectors, and the-state-of-art method in~\cite{thys2019fooling} also generates adversarial patches by minimizing the object score of the detector. Inspired by this, we focus on adversarial face patches to design a baseline method named  \textit{Patch}-\textit{IoU}. Specifically, we firstly define \emph{Adversarial Sample Set} $\{AS_i\}$ as the set of selected samples which are involved in the training optimization of the adversarial patch. Each \emph{Adversarial Sample} $AS_i =(A_i, B_i, S_i, P)$ contains four elements: the anchor $A_i$, the ground-truth bounding box $B_i$, the face confidence score $S_i$ which represents the output of the classification layer with the softmax operation, and the adversarial patch $P$\footnote{All $AS_i$'s share an identical adversarial patch $P$.}, respectively. We freeze all weights of the face detector; the patch $P$ is the only variable to be optimized by gradient ascent algorithm. Our goal is to maximize the following loss function:
\begin{equation} \label{loss}
\setlength\abovedisplayskip{1pt}
\setlength\belowdisplayskip{1pt}
\begin{split}
\mathcal{L}_{Adv}(P) =& -\frac{1}{N}\sum_{i=1}^N \log(S_i),\\
\end{split}
\end{equation}
where $N$ is the size of $\{AS_i\}$.
We use the IoU to select $\{AS_i\}$, that is, each sample in $\{AS_i\}$ should satisfy $\IoU(A_i, B_i)>0.6$, which is exactly the same as the selection of positive samples in the baseline model. Our baseline algorithm \textit{Patch}-\textit{IoU} can be seen from the first row of Figure~\ref{fig:algo}.

\noindent{\textbf{Evaluation details.}}
We follow the same testing settings as that of the SLN (or other face detection framework) baseline model. Similar to the existing works, we set a threshold $\delta=0.99$ to keep high precision: decreasing the scores of the ground-truth faces to be smaller than $\delta$ represents a successful attack.

We show our visualized results in Figure \ref{fig:adv-figure} and the first column of Figure~\ref{fig:noise-init}, i.e., the evolution of the adversarial patch with random initialization. Table~\ref{tab:base-table} (see Baseline, Noise w/o \emph{Patch-IoU} and Noise w/ \emph{Patch-IoU} three lines) presents the corresponding numerical results on precision and recall with (without) \emph{Patch-IoU} optimization. We have three main observations:
\begin{itemize}
\setlength{\itemsep}{1pt}
\setlength{\parsep}{0pt}
\setlength{\parskip}{0pt}
\item 
The drop of recall implies that the detector fails to detect the real faces in the presence of the adversarial patch, i.e., $\Score(\mathrm{Real Face}) < \delta$.
\item
The patch with 100-epoch training appears face-like. The drop of precision implies that the detector \emph{falsely recognizes} the adversarial patch as a human face, i.e., $\Score(\mathrm{Adversarial Patch}) > \delta$.
\item
Attaching an additional face photo to the real faces with the same size and location as the adversarial patch indeed affects precision more than other setups, but we do not obverse significant drop of recall.
\vspace{-0.3cm}
\end{itemize}

\begin{figure*}[t]
\begin{center}
\includegraphics[width=0.95\linewidth]{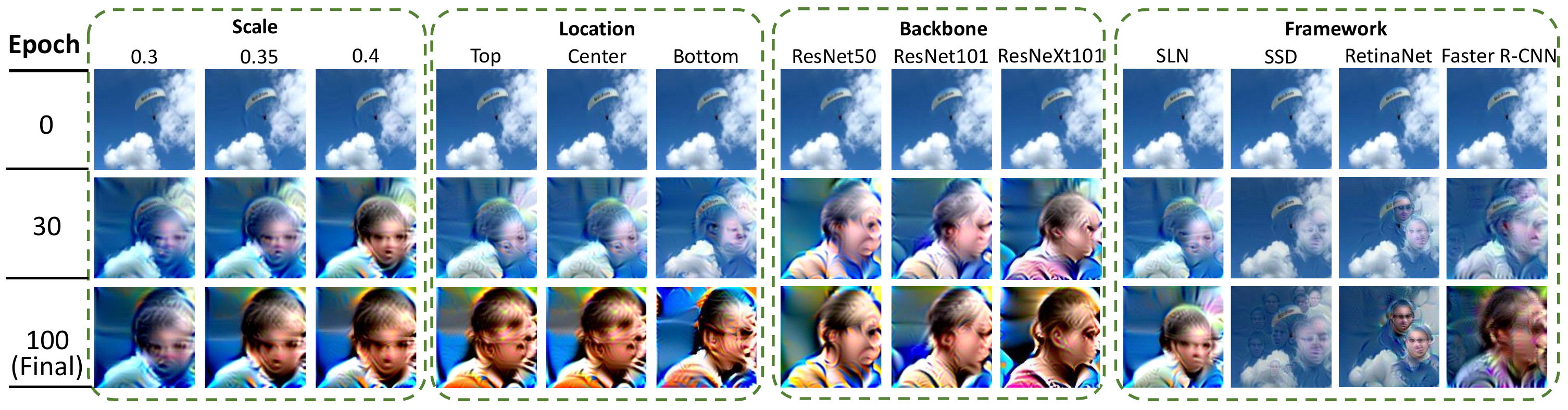}
\end{center}
\vspace{-5ex}
\caption{Optimization results by \textit{Patch}-\textit{IoU} across different scales, locations, backbones and detection frameworks. \emph{Patch-IoU} generates face-like adversarial patch which is falsely detected by various detectors.}
\label{fig:generality}
\vspace{-4ex}
\end{figure*}

\vspace{-0.1cm}
\subsection{Generality}
\vspace{-0.1cm}

The interpretation of adversarial patch is not a unique property of the setup in Section \ref{basic_attack}. Instead, we show that it is a general phenomenon which holds across different initialization, patch locations and scales, backbones, and detection frameworks. 

\noindent{\textbf{Initialization.}}
We randomly select seven images from ImageNet~\cite{imagenet_cvpr09}, three faces from WIDER FACE validation set, and one random image as our initialization. Figure~\ref{fig:noise-init} shows the evolution of patches across different training epochs. We observe that the patches come to resemble human faces, even while initializing the patches with non-face objects or a complex scene.


\noindent{\textbf{Patch locations and scales.}}
To examine whether the interpretation holds across different locations and scales, we run the algorithm with different patch \textit{scales}: $\alpha\in\{0.3, 0.35, 0.4\}$ and \textit{locations}: top $(x^{P}, y^{P}) = (w^i/2, \alpha\sqrt{w^ih^i}/2)$, center $(x^{P}, y^{P}) = (w^i/2, h^i/2)$, and bottom $(x^{P}, y^{P}) = (w^i/2, h^i- \alpha\sqrt{w^ih^i}/2)$. We observe a similar phenomenon for all these setups, as shown in Figure~\ref{fig:generality}.

\noindent{\textbf{Backbones.}}
We see in Figure~\ref{fig:generality} that the adversarial patches look like human faces for different backbones, including ResNet-50, ResNet-101, and ResNext-101.

\noindent{\textbf{Detection frameworks.}}
Besides the face detection framework SLN, we also test three popular detection frameworks: SSD~\cite{liu2016ssd}, RetinaNet~\cite{lin2017focal} and Faster R-CNN~\cite{ren2015faster}. For Faster R-CNN, we use the SLN as our region proposal network, and the RoIAlign~\cite{he2017mask} is applied on each proposal to refine face classification and bounding box regression. Except for the detection architecture, all of the experimental setups for the baseline model and the adversarial patch training are the same. Similarly, we observe that the adversarial patches come to resemble human faces (see Figure~\ref{fig:generality}).



\noindent{\textbf{Numerical results.}}
We also report the numerical results of the algorithm. We set $\delta=0.99$ and show the precision and recall of using various patches to attack the SLN face detector on the WIDER FACE validation set. Table~\ref{tab:base-table} illustrates the effect of eight representative kinds of initialization with (without) \emph{Patch-IoU} optimization. We do not report the numerical results about different patch locations and scales, backbones and detection frameworks, since the results and phenomenon are identical as the initialization. It can be seen that pasting a patch (even initialized as face) without any optimization will cause the recall to drop, but not so drastically. In contrast, the \emph{Patch-IoU} can cause the recall to decrease dramatically across different initialization, leading to a successful attack. However, the adversarial patches also reduce the precision because the face-like patches are falsely detected and the scores of the patches are even higher than those of the true faces. We defer more discussions about evaluation metrics and the issue of precision drop in \emph{Patch-IoU} method to Section~\ref{imporve_opt}.

\noindent{\textbf{Transferability between different frameworks.}} We also study the transferability of adversarial patch between different frameworks. Formally, we attach patches optimized from SSD, RetinaNet and Faster R-CNN, respectively, on each ground-truth bounding box in WIDER FACE validation set, and test their attacking performance on SLN baseline detector. Table~\ref{tab:transferability} shows the numerical results. The patch trained on the Faster R-CNN framework enjoys higher success rate as an attacker on the SLN than the SSD and RetinaNet. 

\begin{table}[t]
\scriptsize
\vspace{-2ex}
  \caption{Precision and recall of different frameworks and transferability of adversarial patch attack from SSD, RetinaNet and Faster R-CNN to SLN under $\delta = 0.99$. A $\rightarrow$ B denotes that the adversarial patch is optimized by detector A and tested on B.}
  \label{tab:transferability}
  \centering
  \vspace{-2ex}
  \begin{tabular}{l|c|c|c|c}
  \hline
   Precision/ Recall & \textit{Easy} & \textit{Medium} & \textit{Hard} & \textit{All} \\
    \hline\hline
    Baseline-SLN & 99.0/ 73.4 & 99.4/ 62.4 & 99.4/ 27.9 & 99.4/ 22.5 \\
    \hline
    \emph{Patch-IoU}-SLN & 2.7/2.7 & 6.5/3.7 & 7.3/1.8 & 7.3/1.4 \\
    \hline
    SSD $\rightarrow$ SLN & 42.5/ 29.9 & 53.4/ 25.1 & 54.1/ 10.7 & 54.1/ 8.6 \\
    RetinaNet $\rightarrow$ SLN & 37.4/ 28.7 & 48.5/ 24.5 & 49.2/ 10.5 & 49.2/ 8.5 \\
    Faster R-CNN $\rightarrow$ SLN & 32.9/ 3.8 & 44.9/ 3.4 & 46.3/ 1.5 & 46.3/ 1.2 \\
    \hline
  \end{tabular}
  \vspace{-5ex}
\end{table}

Besides, we examine generality across training datasets. The final patches are face-like and can be falsely detected by baseline face detector.  To examine the attacking performance of only part of the adversarial patch that is optimized by \textit{Patch-IoU}, we remove a half and one third area of the whole patch and test the performance of the remaining part of the patch on the WIDER FACE validation dataset. Due to limited space, we show these results in Appendix A.

\vspace{-0.1cm}
\subsection{Interpretation of Adversarial Patch.}
\vspace{-0.1cm}

Anchor mechanism with different scales based face detectors provides plenty of facial candidate proposals. This essentially belongs to an ensemble defense strategy compared to classification tasks when adversarial patches strive to lower the classification score of each proposal.  Therefore, the adversarial patch will be optimized towards reducing the scores of more proposals. The previous method \textit{Patch-IoU} only optimizes proposals over a certain range of \textit{IoU}, not including its own adversarial patch. Unconstrained optimization for patches can reduce classification scores of most proposals, yet appearing face-like phenomenon in the face detection task. That makes us rethink what is an effective patch optimization method for deceiving face detectors.

\vspace{-0.2cm}
\section{Improved Optimization of Adversarial Patch}
\vspace{-0.2cm}
\label{imporve_opt}
Current baseline optimization method \textit{Patch}-\textit{IoU} appear face-like phenomenon on the adversarial patch. In this section, we first introduce an evaluation criterion of what kind of patch belongs to a better optimization. Then we propose two improved optimization methods, \textit{Patch}-\textit{Score}-\textit{Focal} and \textit{Patch}-\textit{Combination}, based on above analysis. We demonstrate the effectiveness of the proposed approaches by visualized and numerical results.

\vspace{-0.1cm}
\subsection{Evaluation metric}
\vspace{-0.1cm}

\label{eval_metric}
\noindent{\textbf{Attacking criteria.}}
We set a confidence score threshold $\delta$ to keep high precision ($>0.99$) in order to reduce the possibility of raising false positives. Note that the adversarial patch by \textit{Patch}-\textit{IoU} can be detected as faces (see Section~\ref{basic_attack}). To successfully attack a face detection framework under the policy that none of the bounding boxes in the images should be detected as faces, we define our criteria as follows:
\begin{itemize}
\setlength{\itemsep}{1pt}
\setlength{\parsep}{0pt}
\setlength{\parskip}{0pt}
\item 
\textit{\textbf{Criterion 1}}: Reducing the confidence scores of true faces to be lower than $\delta$;
\item 
\textit{\textbf{Criterion 2}}: Preventing the confidence score of adversarial patch from being higher than $\delta$.
\vspace{-0.2cm}
\end{itemize}
\begin{figure}[t]
\begin{center}
\vspace{-1ex}
\includegraphics[width=0.6\linewidth]{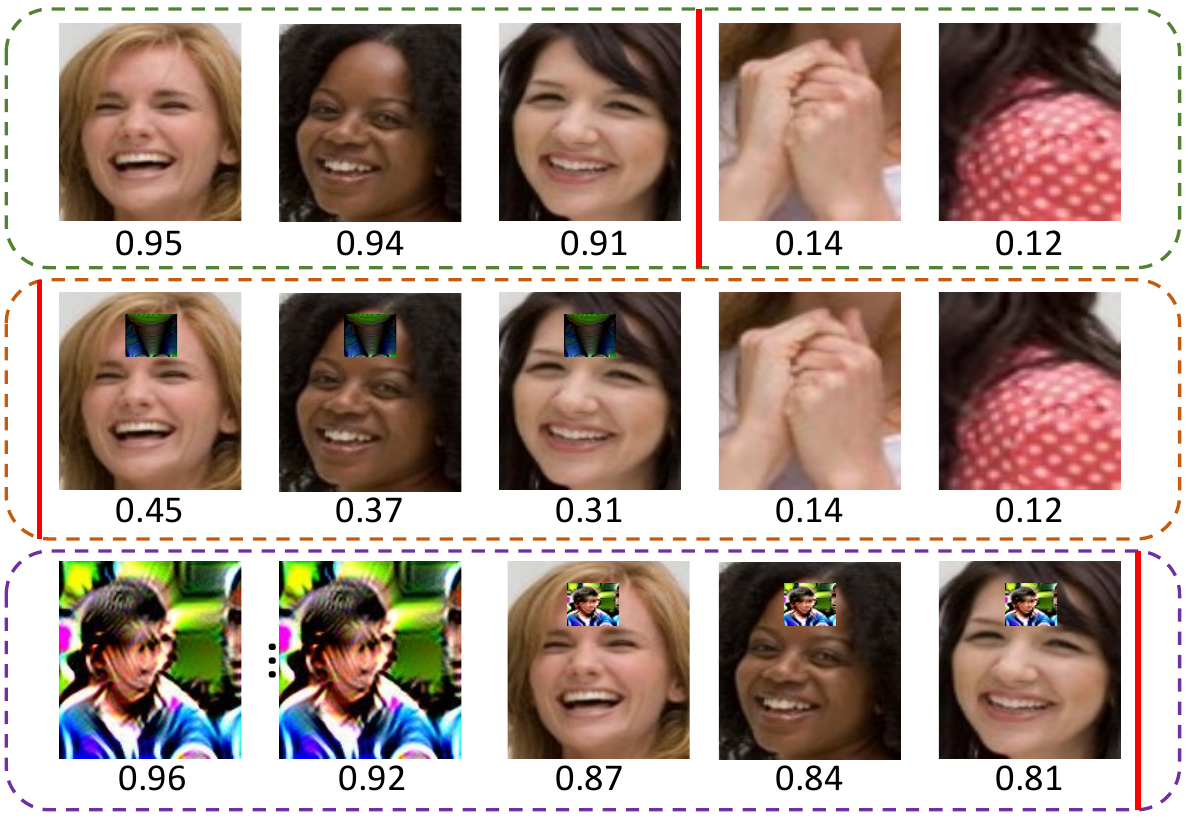}
\end{center}
\vspace{-5ex}
\caption{The first row contains top-5 face proposals for the baseline model and the red line represents the threshold boundary. The second row represents a successful attack. However, the APs in the two rows are the same because the relative rank is the same. The third row illustrates a case where $\Score(\mathrm{Adversarial Patch})>\Score(\mathrm{Face})>\mathrm{Threshold}$. The AP becomes smaller because of the false positives, though the attack fails according to criterion 2. Thus we cannot use AP to evaluate the performances of adversarial patches.} 
\label{fig:reason}
\vspace{-4ex}
\end{figure}

\begin{figure}[t]
\begin{center}
\vspace{-1ex}
\includegraphics[width=0.6\linewidth]{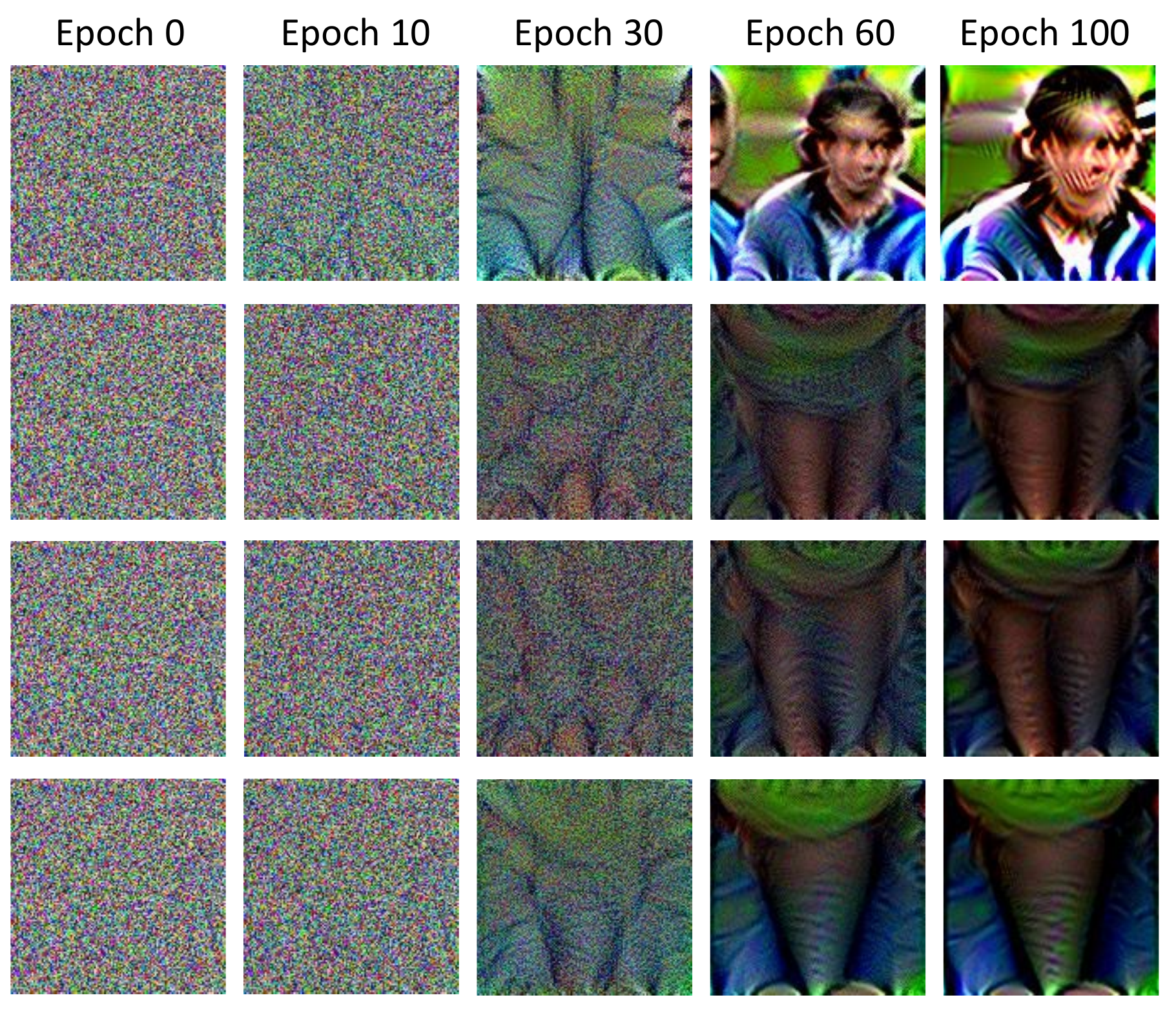}
\end{center}
\vspace{-5ex}
\caption{From top to bottom, the four rows represent the adversarial patches of \textit{Patch}-\textit{IoU}, \textit{Patch}-\textit{Score}, \textit{Patch}-\textit{Score}-\textit{Focal} and \textit{Patch}-\textit{Combination}, respectively.}
\label{fig:opt-patch}
\vspace{-2ex}
\end{figure}

\begin{figure*}[t]
\begin{center}
\includegraphics[width=0.95\linewidth]{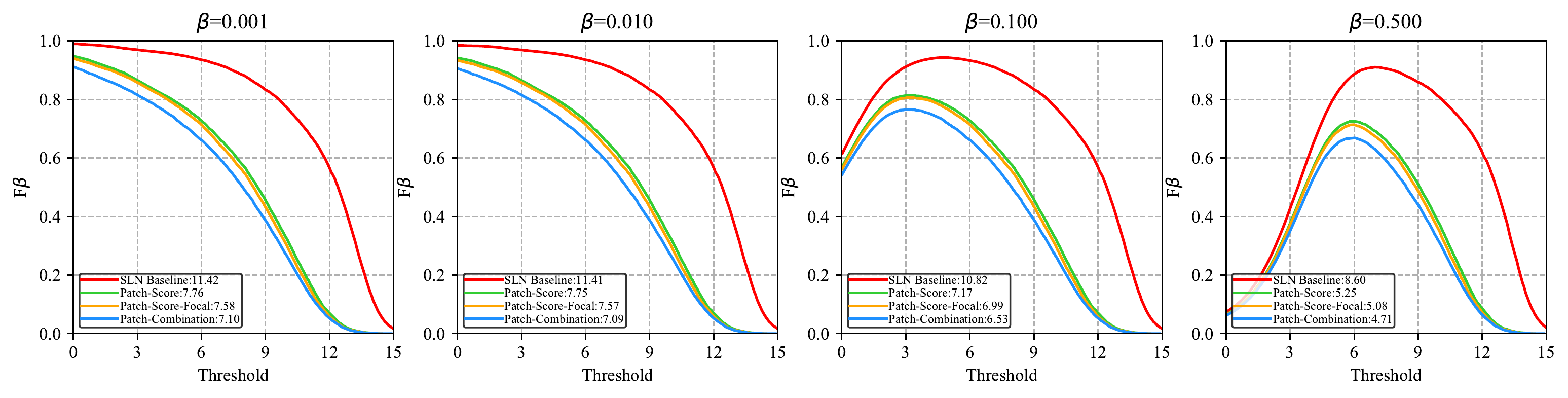}
\end{center}
\vspace{-4ex}
\caption{Threshold-$F_\beta$ curve and $AF_\beta$ under $\beta = [0.001, 0.01, 0.1, 0.5]$. \emph{Patch-Combination} benefits from the confidence scores and location information of ground-truth and the patch and outperforms other methods.}
\label{fig:f-beta}
\vspace{-4ex}
\end{figure*}

\noindent{\textbf{Shortcomings of Average Precision (AP) as an evaluation metric.}}

\begin{itemize}
\setlength{\itemsep}{0pt}
\setlength{\parsep}{0pt}
\setlength{\parskip}{0pt}
\item 
Reducing the confidence scores of true faces does not change the relative rankings among positive (faces) and negative (backgrounds) proposals. As a result, AP remains unchanged even when the attack is successful.
\item
The fake faces appearing on the adversarial patch are treated as false positives. Thus the AP becomes small due to large amounts of false positives. However, this should be considered an unsuccessful attack when the goal is to prevent false positives while suppressing true predictions.
\vspace{-0.3cm}
\end{itemize}
We illustrate the above arguments in Figure~\ref{fig:reason}.

We see that a successful attacking algorithm should reduce the recall of the test images while keeping the precision above the given threshold $\delta$. The observation motivates us to use the recall conditioning on the high precision and the $F_\beta$ score to evaluate the algorithms. $F_\beta$ score is defined as:
\begin{equation*}\label{Fbeta}
\setlength\abovedisplayskip{1.5pt}
\setlength\belowdisplayskip{1.5pt}
  F_\beta = \frac{1 + \beta^2}{\beta^2 \Precision^{-1} + \Recall^{-1}},
\end{equation*} 
where $\beta$ is a hyper-parameter that trades precision off against recall; setting $\beta < 1$ sets more weights to recall and vice versa. We use Average $F_\beta$ ($AF_\beta$), the area under the Threshold-$F_\beta$ curve, to evaluate the attacking algorithms. A lower $AF_\beta$ implies a better attacking algorithm.

\vspace{-0.1cm}
\subsection{Improved optimization}
\vspace{-0.1cm}

As described in Section~\ref{basic_attack}, \textit{Patch-IoU} method may violate \emph{Criterion 2}. We expect the optimized patch to meet two criterions. Therefore, we first introduce a score-based optimization method named \textit{Patch}-\textit{Score}.  Specifically, we set the adversarial sample set $\{AS_i\ = (A_i, B_i, S_i, P)\}$ as those samples with $S_i > \delta - m$, where $m$ is a hyper-parameter on the relaxation of the constraint. This procedure for adversarial sample set selection forces the scores of both adversarial patch and true faces to be lower than predefined threshold $\delta$. We set $\delta - m = 0.5$ as default.

Although \textit{Patch}-\textit{Score} satisfies \emph{Criterion 1} and \emph{Criterion 2} simultaneously, we show that some high-score negative samples may also be selected as adversarial samples $AS_i$, which may degrade the performance as an attacker. In response to this issue, we propose two solutions, namely, \textit{Patch}-\textit{Score}-\textit{Focal} and \textit{Patch}-\textit{Combination}.

\medskip
\noindent{\textbf{\textit{Patch}-\textit{Score}-\textit{Focal} optimization.}} Focal loss~\cite{lin2017focal} aims at solving the extreme imbalance issue between foreground and background proposals in the object detection. The core idea is to assign small weights to the vast majority of easily-classified negatives and prevent them from dominating the classification loss. Our method is inspired from the Focal loss and adapts to the adversarial patch training. Formally, we replace the loss in {\textit{Patch}-\textit{Score}} by
\begin{equation}
\setlength\abovedisplayskip{1pt}
\setlength\belowdisplayskip{1pt}
\mathcal{L}_{Adv}(P) = -\frac{1}{N}\sum_{i=1}^NS_i^{\gamma} \log(S_i),
\end{equation}
where $\gamma$ is a hyper-parameter and $S_i^{\gamma}$ represents the modulating factor which sets different weights for different samples. In contrast to the Focal loss which assigns smaller weights to the easily-classified samples, our goal is to filter out negative proposals whose score are higher than $\delta-m$ and set bigger weights to those negative samples with higher scores. We name this optimization method as \textit{Patch}-\textit{Score}-\textit{Focal} (see the second row in Figure~\ref{fig:algo}). We set $\delta - m = 0.5$ and $\gamma =2$ as suggested in \cite{lin2017focal}.

\medskip
\noindent{\textbf{\textit{Patch}-\textit{Combination} optimization.}}
On one hand, \textit{Patch}-\textit{IoU} aims to select adversarial samples according to the higher IoUs of the ground-truth faces, without any score-related constraints in the adversarial patch optimization. On the other hand, \textit{Patch}-\textit{Score} is to select those samples with confidence scores higher than $\delta - m$, and thus the selected samples may include many negative proposals in the absence of information from the ground-truth faces. We combine the advantages of both methods and propose a new optimization method named \textit{Patch}-\textit{Combination}. Formally, we restrict each adversarial sample $AS_i = (A_i, B_i, S_i, P)$ to satisfy the following conditions: 1) $S_i> \delta - m$; 2) $\IoU(A_i, B_i)>\lambda_1$ or $\IoU(A_i, P)>\lambda_2$. The third row of Figure~\ref{fig:algo} illustrates the methodology. We set $\delta - m = 0.5$, $\lambda_1 = 0.3$ and $\lambda_2 = 0.3$ as default.

\begin{table}[t]
\scriptsize
  \caption{Precision and recall comparisons of Baseline-SLN, \textit{Patch}-\textit{Score}, \textit{Patch}-\textit{Score}-\textit{Focal} and \textit{Patch}-\textit{Combination} with $\delta = 0.99$.}
  \label{tab:thrs-table}
  \centering
  \begin{tabular}{l|c|c|c|c}
    \hline
   Precision/ Recall & \emph{Easy} & \emph{Medium} & \emph{Hard} & \emph{All} \\
    \hline\hline
    Baseline-SLN & 99.0 / 73.4 & 99.4 / 62.4 & 99.4 / 27.9 & 99.4 / 22.5 \\

    \textit{Patch}-\textit{Score} & 98.5 / 25.5 & 98.9 / 19.7 & 99.0 / 8.3 & 99.0 / 6.7 \\
    \textit{Patch}-\textit{Score}-\textit{Focal} & 98.4 / 23.1 & 98.9 / 17.9 & 98.9 / 7.6 & 98.9 / 6.1 \\
    \textit{Patch}-\textit{Combination} & 98.2 / \textbf{20.6} & 98.7 / \textbf{15.6} & 98.7 / \textbf{6.6} & 98.7 / \textbf{5.4} \\
    \hline
  \end{tabular}
 \vspace{-4ex}
\end{table}

\vspace{-0.1cm}
\subsection{Experimental results}
\vspace{-0.1cm}

We use WIDER FACE training set and SLN baseline model for adversarial patch training; except for the adversarial sample set selection procedure, the same optimization setups for \emph{Patch-IoU} training are applied to the \emph{Patch-Score}, \emph{Patch-Score-Focal} and \emph{Patch-Combination} methods. We use WIDER FACE validation set to evaluate the algorithms (see the second to the fourth rows of Figure~\ref{fig:opt-patch} for visualized results). In contrast to \textit{Patch}-\textit{IoU} method, no faces can be detected by our improved optimization algorithms since they all satisfy \emph{Criterion 1} and \emph{Criterion 2}.

Besides the visualized results, we also show numerical results. Figure~\ref{fig:f-beta} shows four Threshold-$F_\beta$\footnote{Define $s_p$ and $s_n$ as positive and negative logits. We compute $s = s_p - s_n$ as the confidence score when plotting Threshold-$F_\beta$ curve for better visualization.} curves and $AF_\beta$ (lower $AF_\beta$ means a better attack) under different $\beta = [0.001, 0.01, 0.1, 0.5]$. Table~\ref{tab:thrs-table} also shows the comparisons of precision and recall with $\delta = 0.99$. Moreover,  we also reduce the adversarial patch to different proportions and paste the patch on different positions for comparisons in Appendix B. In Appendix C, we examine the transferability of the adversarial patch between different models.  \textit{Patch}-\textit{Combination} and \textit{Patch}-\textit{Score}-\textit{Focal} achieve better performance than \textit{Patch}-\textit{Score} on extensive experiments, with better optimization design of adversarial sample set. This is because \textit{Patch}-\textit{Combination} benefits from the interpretation  that attacking an ensemble defense (\emph{e.g.} plenty of proposals in face detection) should fully consider the full use of global information in the optimization. 


\vspace{-0.3cm}
\section{Conclusions}
\vspace{-0.2cm}
In this paper, we perform a comprehensive interpretation of adversarial patches to state-of-the-art anchor based face detectors. Firstly we show a face-like phenomenon of the generated adversarial patches by previous method \emph{Patch-IoU}, which makes the detectors falsely recognize the patches as human faces across different initialization, locations and scales, backbones, etc. That is very instructive for the understanding of universal adversarial
samples from an optimization perspective. Besides, we propose \emph{Patch-Score-Focal} and \emph{Patch-Combination} methods to obtain more effective adversarial patches. Extensive experiments verify the effectiveness of the proposed methods across different scales, transferability, etc. We also believe that these promising insights and methods will inspire further studies for the community. 

\medskip
\noindent{\textbf{Acknowledgement.}}
We thank Gregory Shakhnarovich for helping to improve the writing of this paper and valuable suggestions on the experimental designs. X. Yang and J. Zhu  were supported by the National Key Research and Development Program of China (No. 2017YFA0700904), NSFC Projects (Nos. 61620106010, U19B2034, U181146), Beijing Academy of Artificial Intelligence, Tsinghua-Huawei Joint Research Program, Tiangong Institute for Intelligent Computing, and NVIDIA NVAIL Program with GPU/DGX Acceleration. H. Zhang was supported in part by the Defense Advanced Research Projects Agency under cooperative agreement HR00112020003.

\clearpage
%
%

\appendix
\section{Generality Experiments}
\begin{figure}[!htp]
\begin{center}
\includegraphics[width=0.5\linewidth]{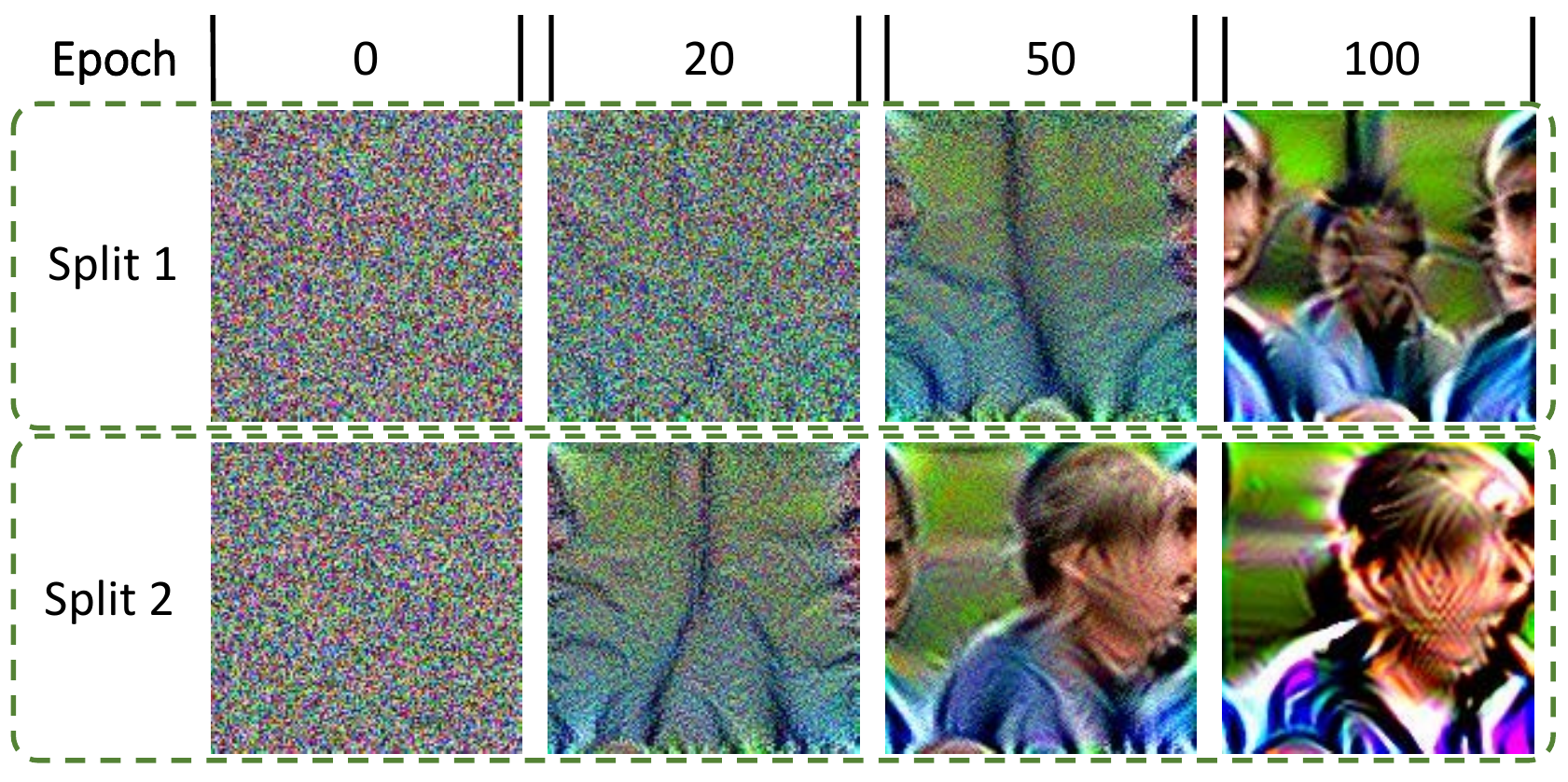}
\end{center}
\vspace{-2ex}
\caption{Optimization results by \textit{Patch-IoU} on WIDER FACE training subset, split 1 and split 2. \textit{Patch-IoU} generates face-like adversarial patch which is falsely detected by baseline face detector.}
\label{fig:split}
\end{figure}

\noindent{\textbf{Generality of training dataset.}} We randomly split WIDER FACE training dataset into two subsets, namely, split 1 and split 2, to verify the generality of different training sources. Figure~\ref{fig:split} shows the evolution of adversarial samples opAs illustrated in Table~\ref{tab:scales}, we reduce the patch to different proportions from 90\% to 60\%. As the scale of patches decreases, there exists a general downward trend for performance. We also paste adversarial patches on different 5 locations from top to bottom for comparisons in Table~\ref{tab:locations}. The results show that it is effective to stick on different locations.

As illustrated in Table~\ref{tab:scales}, we reduce the patch to different proportions from 90\% to 60\%. As the scale of patches decreases, there exists a general downward trend for performance. We also paste adversarial patches on different 5 locations from top to bottom for comparisons in Table~\ref{tab:locations}. The results show that it is effective to stick on different locations.

timized by \textit{Patch-IoU} method on each split. The final patches are face-like and can be falsely detected by baseline face detector. The two patches look slightly different, partially because of the non-convex properties of optimization~\cite{zhang2019deep,zhang2018stackelberg} and different initialization.

\noindent{\textbf{Attack by part of patch.}} To examine the attacking performance of part of the adversarial patch that is optimized by \textit{Patch-IoU}, we remove a half and one third area of the whole patch and test the performance of the remaining part of the patch on the WIDER FACE validation dataset. Table~\ref{tab:cut} shows the associated numerical results. We see that removing a part of the patch hurts the performance of the adversarial patch as an attacker.
\begin{table}[t]
  \caption{Attacking performance of parts of adversarial patches. Half (One-third)-\textit{Part} means removing one half (one-third) area of the whole patches.}
  \label{tab:cut}
  \centering
  \begin{tabular}{l|c|c|c|c}
    \hline
   Precision/ Recall & \textit{Easy} & \textit{Medium} & \textit{Hard} & \textit{All} \\
    \hline\hline
    Baseline-SLN & 99.0/ 73.4 & 99.4/ 62.4 & 99.4/ 27.9 & 99.4/ 22.5 \\
    \hline
    \emph{Patch-IoU}-SLN & 2.7/2.7 & 6.5/3.7 & 7.3/1.8 & 7.3/1.4 \\
    \hline
    Half-\textit{Top} & 93.3/ 38.8 & 95.9/ 35.6 & 96.1/ 15.6 & 96.1/ 12.5 \\
    Half-\textit{Bottom} & 99.2/ 47.4 & 99.5/ 40.1 & 99.5/17.2 & 99.5/13.8 \\
    Half-\textit{Left} & 98.5/ 41.2 & 99.1/ 36.3 & 99.1/15.8 & 99.1/ 12.7 \\
    Half-\textit{Right} & 82.0/ 30.6 & 88.8/28.7 & 89.2/12.5 & 89.2/ 10.1 \\
    \hline
    One-third-\textit{Top} & 25.3/18.7 & 39.1/19.1 & 40.6/8.5 & 40.6/6.8 \\
    One-third-\textit{Bottom} &  74.6/23.0 & 83.2/12.0 & 83.7/9.1 & 83.7/7.3\\
    One-third-\textit{Left} & 88.1/24.1 & 92.9/22.8 & 93.2/10.1 & 93.2/8.1 \\
    One-third-\textit{Right} & 12.8/11.9 & 22.44/12.7 & 23.6/5.7 & 23.6/4.6 \\
    \hline
  \end{tabular}
\end{table}

\section{Improved Patches on Different Scales and Locations}

As illustrated in Table~\ref{tab:scales}, we reduce the patch to different proportions from 90\% to 60\%. As the scale of patches decreases, there exists a general downward trend for performance. We also paste adversarial patches on different 5 locations from top to bottom for comparisons in Table~\ref{tab:locations}. The results show that it is effective to stick on different locations.

\begin{table}[t]
\small
\setlength{\tabcolsep}{2mm}
  \caption{Precision and recall comparisons of  \textit{Patch}-\textit{Score}, \textit{Patch}-\textit{Score}-\textit{Focal} and \textit{Patch}-\textit{Combination} based on different scales with $\delta = 0.99$.}
  \label{tab:scales}
  \centering
  \begin{tabular}{c|c|c|c|c|c}
    \hline
  Scale & Precision/ Recall & \emph{Easy} & \emph{Medium} & \emph{Hard} & \emph{All} \\
    \hline\hline
      90\% & \tabincell{c}{\textit{Patch}-\textit{Score} \\ \textit{Patch}-\textit{Score}-\textit{Focal} \\ \textit{Patch}-\textit{Combination}}  &
      \tabincell{c}{98.9/40.0 \\ 98.9/37.7 \\ 98.9/35.4} &
      \tabincell{c}{99.3/30.7 \\ 99.2/29.5 \\ 99.2/27.0} &
      \tabincell{c}{99.2/13.0 \\ 99.2/12.5 \\ 99.2/11.5} &
      \tabincell{c}{99.3/10.5 \\ 99.2/10.0 \\ 99.2/9.2} \\
      
      \hline
      80\% & \tabincell{c}{\textit{Patch}-\textit{Score} \\ \textit{Patch}-\textit{Score}-\textit{Focal} \\ \textit{Patch}-\textit{Combination}}  &
      \tabincell{c}{99.1/50.1 \\ 99.1/49.2 \\ 99.1/46.7} &
      \tabincell{c}{99.4/39.7 \\ 99.4/39.1 \\ 99.4/36.1} &
      \tabincell{c}{99.4/16.9 \\ 99.4/16.8 \\ 99.4/15.4} &
      \tabincell{c}{99.4/13.6 \\ 99.4/13.3 \\ 99.4/12.4} \\

      \hline
      70\% & \tabincell{c}{\textit{Patch}-\textit{Score} \\ \textit{Patch}-\textit{Score}-\textit{Focal} \\ \textit{Patch}-\textit{Combination}}  &
      \tabincell{c}{99.2/58.7 \\ 99.2/58.2 \\ 99.2/55.9} &
      \tabincell{c}{99.5/47.0 \\ 99.5/46.6 \\ 99.4/44.3} &
      \tabincell{c}{99.5/20.0 \\ 99.5/19.8 \\ 99.4/18.9} &
      \tabincell{c}{99.5/16.1 \\ 99.5/15.9 \\ 99.4/15.2} \\
      \hline
      60\% & \tabincell{c}{\textit{Patch}-\textit{Score} \\ \textit{Patch}-\textit{Score}-\textit{Focal} \\ \textit{Patch}-\textit{Combination}}  &
      \tabincell{c}{99.1/64.2 \\ 99.1/63.9 \\ 99.1/62.5} &
      \tabincell{c}{99.4/52.0 \\ 99.4/51.8 \\ 99.4/50.2} &
      \tabincell{c}{99.4/22.2 \\ 99.4/22.1 \\ 99.4/21.4} &
      \tabincell{c}{99.4/17.9 \\ 99.4/17.8 \\ 99.4/17.2} \\
      
    \hline
  \end{tabular}
  \vspace{-0.5cm}
\end{table}

\begin{table}[t]
\small
  \caption{Precision and recall comparisons of  \textit{Patch}-\textit{Score}, \textit{Patch}-\textit{Score}-\textit{Focal} and \textit{Patch}-\textit{Combination} based on different locations with $\delta = 0.99$.}
  \label{tab:locations}
  \centering
  \begin{tabular}{c|c|c|c|c|c}
    \hline
  Location & Precision/ Recall & \emph{Easy} & \emph{Medium} & \emph{Hard} & \emph{All} \\
    \hline\hline
      Top & \tabincell{c}{\textit{Patch}-\textit{Score} \\ \textit{Patch}-\textit{Score}-\textit{Focal} \\ \textit{Patch}-\textit{Combination}}  &
      \tabincell{c}{98.5/25.9 \\ 98.4/23.4 \\ 98.2/20.6} &
      \tabincell{c}{99.0/20.0 \\ 98.9/18.1 \\ 98.7/15.7} &
      \tabincell{c}{99.0/8.5 \\ 98.9/7.7 \\ 98.8/6.7} &
      \tabincell{c}{99.0/6.8 \\ 98.9/6.2 \\ 98.8/5.3} \\
      \hline
      \tabincell{c}{Center of \\ top and middle} & \tabincell{c}{\textit{Patch}-\textit{Score} \\ \textit{Patch}-\textit{Score}-\textit{Focal} \\ \textit{Patch}-\textit{Combination}}  &
      \tabincell{c}{97.9/17.1 \\ 97.7/15.1 \\ 97.7/15.3} &
      \tabincell{c}{98.7/14.8 \\ 98.6/12.9 \\ 98.6/13.0} &
      \tabincell{c}{98.7/6.3 \\ 98.6/5.5 \\ 98.6/5.6} &
      \tabincell{c}{98.7/5.1 \\ 98.6/4.5 \\ 98.6/4.5} \\
      
      \hline
      Middle & \tabincell{c}{\textit{Patch}-\textit{Score} \\ \textit{Patch}-\textit{Score}-\textit{Focal} \\ \textit{Patch}-\textit{Combination}}  &
      \tabincell{c}{97.9/17.9 \\ 97.5/14.6 \\ 97.5/14.8} &
      \tabincell{c}{98.6/15.0 \\ 98.3/12.1 \\ 98.3/11.8} &
      \tabincell{c}{98.7/6.1 \\ 98.4/5.2 \\ 98.4/5.0} &
      \tabincell{c}{98.7/5.2 \\ 98.4/4.2 \\ 98.4/4.0} \\
      \hline
      \tabincell{c}{Center of \\ bottom and middle} & \tabincell{c}{\textit{Patch}-\textit{Score} \\ \textit{Patch}-\textit{Score}-\textit{Focal} \\ \textit{Patch}-\textit{Combination}}  &
      \tabincell{c}{98.1/24.1 \\ 98.4/25.1 \\ 98.3/23.3} &
      \tabincell{c}{98.7/18.5 \\ 98.8/18.3 \\ 98.7/15.8} &
      \tabincell{c}{98.7/7.8 \\ 98.8/7.8 \\ 98.7/6.7} &
      \tabincell{c}{98.7/6.3 \\ 98.8/6.2 \\ 98.7/5.4} \\
      \hline 
      Bottom & \tabincell{c}{\textit{Patch}-\textit{Score} \\ \textit{Patch}-\textit{Score}-\textit{Focal} \\ \textit{Patch}-\textit{Combination}}  &
      \tabincell{c}{98.9/39.8 \\ 98.8/38.8 \\ 98.8/39.0} &
      \tabincell{c}{99.3/31.1 \\ 99.2/31.2 \\ 99.2/28.8} &
      \tabincell{c}{99.3/13.3 \\ 99.2/13.3 \\ 99.2/12.1} &
      \tabincell{c}{98.3/10.7 \\ 98.8/10.7 \\ 98.7/9.7} \\
      
    \hline
  \end{tabular}
\end{table}

\section{Transferablility on Different Models}

We examine the transferability of adversarial patches between different models in Table~\ref{tab:trans}, including ResNet-50, ResNet-101 and ResNext-101. ResNet-18 is a surrogate white-box model. It is also observed
that proposed adversarial patches have some effects on unknown black-box models.

\begin{table}[t]
\small
\setlength{\tabcolsep}{2mm}
  \caption{Precision and recall of  \textit{Patch}-\textit{Score}-\textit{Focal} and \textit{Patch}-\textit{Combination} based on different models with $\delta = 0.99$.}
  \label{tab:trans}
  \centering
  \begin{tabular}{c|c|c|c|c|c}
    \hline
  Models & Precision/ Recall & \emph{Easy} & \emph{Medium} & \emph{Hard} & \emph{All} \\
    \hline\hline
      ResNet-50 & \tabincell{c}{Baseline \\ \textit{Patch}-\textit{Score}-\textit{Focal} \\ \textit{Patch}-\textit{Combination}}  &
      \tabincell{c}{99.0/81.9 \\ 98.8/46.1 \\ 98.9/47.6} &
      \tabincell{c}{99.4/75.2 \\ 99.2/36.0 \\ 99.2/37.2} &
      \tabincell{c}{99.6/40.3 \\ 99.2/15.4 \\ 99.2/15.9} &
      \tabincell{c}{99.6/32.5 \\ 99.2/12.4 \\ 99.2/12.8} \\
      \hline
      
      ResNet-101 & \tabincell{c}{Baseline \\ \textit{Patch}-\textit{Score}-\textit{Focal} \\ \textit{Patch}-\textit{Combination}}  &
      \tabincell{c}{99.0/83.6 \\ 98.9/54.5 \\ 99.0/56.0} &
      \tabincell{c}{99.4/77.1 \\ 99.3/45.2 \\ 99.3/46.7} &
      \tabincell{c}{99.6/39.8 \\ 99.3/19.3 \\ 99.4/20.0} &
      \tabincell{c}{99.6/32.1 \\ 99.3/15.6 \\ 99.4/16.1} \\
      \hline
      ResNext-101 & \tabincell{c}{Baseline \\ \textit{Patch}-\textit{Score}-\textit{Focal} \\ \textit{Patch}-\textit{Combination}}  &
      \tabincell{c}{99.0/82.7 \\ 98.9/54.6 \\ 98.3/55.7} &
      \tabincell{c}{99.4/75.6 \\ 99.2/42.3 \\ 98.7/43.4} &
      \tabincell{c}{99.6/41.0 \\ 99.3/17.9 \\ 98.7/18.4} &
      \tabincell{c}{99.6/33.0 \\ 99.3/14.2 \\ 98.7/14.8} \\
      \hline 
      
  \end{tabular}
  \vspace{-0.5cm}
\end{table}

\bibliographystyle{splncs04}
\bibliography{egbib}

\begin{thebibliography}{10}
\providecommand{\url}[1]{\texttt{#1}}
\providecommand{\urlprefix}{URL }
\providecommand{\doi}[1]{https://doi.org/#1}

\bibitem{athalye2017synthesizing}
Athalye, A., Engstrom, L., Ilyas, A., Kwok, K.: Synthesizing robust adversarial
  examples. arXiv preprint arXiv:1707.07397  (2017)

\bibitem{biggio2013evasion}
Biggio, B., Corona, I., Maiorca, D., Nelson, B., {\v{S}}rndi{\'c}, N., Laskov,
  P., Giacinto, G., Roli, F.: Evasion attacks against machine learning at test
  time. In: Joint European conference on machine learning and knowledge
  discovery in databases. pp. 387--402 (2013)

\bibitem{blum2020random}
Blum, A., Dick, T., Manoj, N., Zhang, H.: Random smoothing might be unable to
  certify $\ell_{\infty}$ robustness for high-dimensional images. arXiv
  preprint arXiv:2002.03517  \textbf{2}(2) (2020)

\bibitem{brendel2020adversarial}
Brendel, W., Rauber, J., Kurakin, A., Papernot, N., Veliqi, B., Mohanty, S.P.,
  Laurent, F., Salath{\'e}, M., Bethge, M., Yu, Y., et~al.: Adversarial vision
  challenge. In: The NeurIPS'18 Competition, pp. 129--153 (2020)

\bibitem{brown2017adversarial}
Brown, T.B., Man{\'e}, D., Roy, A., Abadi, M., Gilmer, J.: Adversarial patch.
  arXiv preprint arXiv:1712.09665  (2017)

\bibitem{chen2018shapeshifter}
Chen, S.T., Cornelius, C., Martin, J., Chau, D.H.P.: Shapeshifter: Robust
  physical adversarial attack on faster r-cnn object detector. In: Joint
  European Conference on Machine Learning and Knowledge Discovery in Databases.
  pp. 52--68 (2018)

\bibitem{imagenet_cvpr09}
Deng, J., Dong, W., Socher, R., Li, L.J., Li, K., Fei-Fei, L.: {ImageNet: A
  Large-Scale Hierarchical Image Database}. In: CVPR09 (2009)

\bibitem{dong2020benchmarking}
Dong, Y., Fu, Q.A., Yang, X., Pang, T., Su, H., Xiao, Z., Zhu, J.: Benchmarking
  adversarial robustness on image classification. In: Proceedings of the
  IEEE/CVF Conference on Computer Vision and Pattern Recognition (CVPR). pp.
  321--331 (2020)

\bibitem{eykholt2018physical}
Eykholt, K., Evtimov, I., Fernandes, E., Li, B., Rahmati, A., Tramer, F.,
  Prakash, A., Kohno, T., Song, D.: Physical adversarial examples for object
  detectors. arXiv preprint arXiv:1807.07769  (2018)

\bibitem{eykholt2018robust}
Eykholt, K., Evtimov, I., Fernandes, E., Li, B., Rahmati, A., Xiao, C.,
  Prakash, A., Kohno, T., Song, D.: Robust physical-world attacks on deep
  learning visual classification. In: IEEE Conference on Computer Vision and
  Pattern Recognition. pp. 1625--1634 (2018)

\bibitem{eykholt2017note}
Eykholt, K., Evtimov, I., Fernandes, E., Li, B., Song, D., Kohno, T., Rahmati,
  A., Prakash, A., Tramer, F.: Note on attacking object detectors with
  adversarial stickers. arXiv preprint arXiv:1712.08062  (2017)

\bibitem{goodfellow2014explaining}
Goodfellow, I.J., Shlens, J., Szegedy, C.: Explaining and harnessing
  adversarial examples. In: International Conference on Learning
  Representations (ICLR) (2015)

\bibitem{he2017mask}
He, K., Gkioxari, G., Doll{\'a}r, P., Girshick, R.: Mask r-cnn. In: Proceedings
  of the IEEE international conference on computer vision. pp. 2961--2969
  (2017)

\bibitem{he2016deep}
He, K., Zhang, X., Ren, S., Sun, J.: Deep residual learning for image
  recognition. In: Proceedings of the IEEE conference on computer vision and
  pattern recognition. pp. 770--778 (2016)

\bibitem{jain2010fddb}
Jain, V., Learned-Miller, E.: Fddb: A benchmark for face detection in
  unconstrained settings  (2010)

\bibitem{jia2017adversarial}
Jia, R., Liang, P.: Adversarial examples for evaluating reading comprehension
  systems. arXiv preprint arXiv:1707.07328  (2017)

\bibitem{komkov2019advhat}
Komkov, S., Petiushko, A.: Advhat: Real-world adversarial attack on arcface
  face id system. arXiv preprint arXiv:1908.08705  (2019)

\bibitem{Kurakin2016}
Kurakin, A., Goodfellow, I., Bengio, S.: Adversarial examples in the physical
  world. In: International Conference on Learning Representations (ICLR)
  Workshops (2017)

\bibitem{lee2019physical}
Lee, M., Kolter, Z.: On physical adversarial patches for object detection.
  arXiv preprint arXiv:1906.11897  (2019)

\bibitem{li2019dsfd}
Li, J., Wang, Y., Wang, C., Tai, Y., Qian, J., Yang, J., Wang, C., Li, J.,
  Huang, F.: Dsfd: dual shot face detector. In: Proceedings of the IEEE
  Conference on Computer Vision and Pattern Recognition. pp. 5060--5069 (2019)

\bibitem{li2019hiding}
Li, Y., Yang, X., Wu, B., Lyu, S.: Hiding faces in plain sight: Disrupting ai
  face synthesis with adversarial perturbations. arXiv preprint
  arXiv:1906.09288  (2019)

\bibitem{lin2017feature}
Lin, T.Y., Doll{\'a}r, P., Girshick, R., He, K., Hariharan, B., Belongie, S.:
  Feature pyramid networks for object detection. In: Proceedings of the IEEE
  conference on computer vision and pattern recognition. pp. 2117--2125 (2017)

\bibitem{lin2017focal}
Lin, T.Y., Goyal, P., Girshick, R., He, K., Doll{\'a}r, P.: Focal loss for
  dense object detection. In: Proceedings of the IEEE international conference
  on computer vision. pp. 2980--2988 (2017)

\bibitem{liu2016ssd}
Liu, W., Anguelov, D., Erhan, D., Szegedy, C., Reed, S., Fu, C.Y., Berg, A.C.:
  Ssd: Single shot multibox detector. In: European conference on computer
  vision. pp. 21--37. Springer (2016)

\bibitem{liu2018dpatch}
Liu, X., Yang, H., Liu, Z., Song, L., Li, H., Chen, Y.: Dpatch: An adversarial
  patch attack on object detectors. arXiv preprint arXiv:1806.02299  (2018)

\bibitem{madry2017towards}
Madry, A., Makelov, A., Schmidt, L., Tsipras, D., Vladu, A.: Towards deep
  learning models resistant to adversarial attacks. In: International
  Conference on Learning Representations (ICLR) (2018)

\bibitem{ming2019group}
Ming, X., Wei, F., Zhang, T., Chen, D., Wen, F.: Group sampling for scale
  invariant face detection. In: Proceedings of the IEEE Conference on Computer
  Vision and Pattern Recognition. pp. 3446--3456 (2019)

\bibitem{Nguyen2015}
Nguyen, A., Yosinski, J., Clune, J.: Deep neural networks are easily fooled:
  High confidence predictions for unrecognizable images. In: The IEEE
  Conference on Computer Vision and Pattern Recognition (CVPR). pp. 427--436
  (2015)

\bibitem{PapernotDistillation2016}
Papernot, N., McDaniel, P., Wu, X., Jha, S., Swami, A.: Distillation as a
  defense to adversarial perturbations against deep neural networks. In: IEEE
  Symposium on Security and Privacy (2016)

\bibitem{ren2015faster}
Ren, S., He, K., Girshick, R., Sun, J.: Faster r-cnn: Towards real-time object
  detection with region proposal networks. In: Advances in neural information
  processing systems. pp. 91--99 (2015)

\bibitem{sharif2016accessorize}
Sharif, M., Bhagavatula, S., Bauer, L., Reiter, M.K.: Accessorize to a crime:
  Real and stealthy attacks on state-of-the-art face recognition. In:
  Proceedings of the 2016 ACM SIGSAC Conference on Computer and Communications
  Security. pp. 1528--1540 (2016)

\bibitem{sharif2019general}
Sharif, M., Bhagavatula, S., Bauer, L., Reiter, M.K.: A general framework for
  adversarial examples with objectives. ACM Transactions on Privacy and
  Security (TOPS)  \textbf{22}(3), ~16 (2019)

\bibitem{szegedy2013intriguing}
Szegedy, C., Zaremba, W., Sutskever, I., Bruna, J., Erhan, D., Goodfellow, I.,
  Fergus, R.: Intriguing properties of neural networks. In: International
  Conference on Learning Representations (ICLR) (2014)

\bibitem{thys2019fooling}
Thys, S., Van~Ranst, W., Goedem{\'e}, T.: Fooling automated surveillance
  cameras: adversarial patches to attack person detection. In: Proceedings of
  the IEEE Conference on Computer Vision and Pattern Recognition Workshops
  (2019)

\bibitem{tsipras2018robustness}
Tsipras, D., Santurkar, S., Engstrom, L., Turner, A., Madry, A.: Robustness may
  be at odds with accuracy. arXiv preprint arXiv:1805.12152  (2018)

\bibitem{wei2020point}
Wei, F., Sun, X., Li, H., Wang, J., Lin, S.: Point-set anchors for object
  detection, instance segmentation and pose estimation. arXiv preprint
  arXiv:2007.02846  (2020)

\bibitem{xie2017adversarial}
Xie, C., Wang, J., Zhang, Z., Zhou, Y., Xie, L., Yuille, A.: Adversarial
  examples for semantic segmentation and object detection. In: IEEE
  International Conference on Computer Vision. pp. 1369--1378 (2017)

\bibitem{yamada2013privacy}
Yamada, T., Gohshi, S., Echizen, I.: Privacy visor: Method for preventing face
  image detection by using differences in human and device sensitivity. In:
  IFIP International Conference on Communications and Multimedia Security. pp.
  152--161 (2013)

\bibitem{yang2016wider}
Yang, S., Luo, P., Loy, C.C., Tang, X.: Wider face: A face detection benchmark.
  In: IEEE Conference on Computer Vision and Pattern Recognition (CVPR) (2016)

\bibitem{yang2020adversarial}
Yang, Y.Y., Rashtchian, C., Zhang, H., Salakhutdinov, R., Chaudhuri, K.:
  Adversarial robustness through local lipschitzness. arXiv preprint
  arXiv:2003.02460  (2020)

\bibitem{you2020greedynas}
You, S., Huang, T., Yang, M., Wang, F., Qian, C., Zhang, C.: Greedynas: Towards
  fast one-shot nas with greedy supernet. In: Proceedings of the IEEE/CVF
  Conference on Computer Vision and Pattern Recognition. pp. 1999--2008 (2020)

\bibitem{you2017learning}
You, S., Xu, C., Xu, C., Tao, D.: Learning from multiple teacher networks. In:
  Proceedings of the 23rd ACM SIGKDD International Conference on Knowledge
  Discovery and Data Mining. pp. 1285--1294 (2017)

\bibitem{zhang2019towards}
Zhang, H., Wang, J.: Towards adversarially robust object detection. In: IEEE
  International Conference on Computer Vision. pp. 421--430 (2019)

\bibitem{zhang2019deep}
Zhang, H., Shao, J., Salakhutdinov, R.: Deep neural networks with multi-branch
  architectures are intrinsically less non-convex. In: International Conference
  on Artificial Intelligence and Statistics. pp. 1099--1109 (2019)

\bibitem{zhang2018stackelberg}
Zhang, H., Xu, S., Jiao, J., Xie, P., Salakhutdinov, R., Xing, E.P.:
  Stackelberg gan: Towards provable minimax equilibrium via multi-generator
  architectures. arXiv preprint arXiv:1811.08010  (2018)

\bibitem{zhang2019theoretically}
Zhang, H., Yu, Y., Jiao, J., Xing, E.P., Ghaoui, L.E., Jordan, M.I.:
  Theoretically principled trade-off between robustness and accuracy. In:
  International Conference on Machine Learning (ICML) (2019)

\bibitem{zhang2019single}
Zhang, S., Wen, L., Shi, H., Lei, Z., Lyu, S., Li, S.Z.: Single-shot
  scale-aware network for real-time face detection. International Journal of
  Computer Vision  \textbf{127}(6-7),  537--559 (2019)

\bibitem{zhu2018seeing}
Zhu, C., Tao, R., Luu, K., Savvides, M.: Seeing small faces from robust
  anchor's perspective. In: Proceedings of the IEEE Conference on Computer
  Vision and Pattern Recognition. pp. 5127--5136 (2018)

\bibitem{zitnick2014edge}
Zitnick, C.L., Doll{\'a}r, P.: Edge boxes: Locating object proposals from
  edges. In: European conference on computer vision. pp. 391--405. Springer
  (2014)

\end{thebibliography}
\end{document}